\documentclass[letterpaper, 10 pt, conference]{ieeeconf}  

\IEEEoverridecommandlockouts                              

\usepackage{cite}
\usepackage{lipsum}
\usepackage[inkscapelatex=false]{svg}
\usepackage{amssymb,amsfonts}
\usepackage[fleqn]{amsmath}
\usepackage{algorithmic}
\usepackage{booktabs}
\usepackage{graphicx}
\usepackage{pifont}
\usepackage{textcomp}
\usepackage{xcolor}
\usepackage{array}
\usepackage{cellspace}
\usepackage{array}
\usepackage{colortbl} 
\usepackage{multirow}
\usepackage{subcaption}
\usepackage{color,soul}
\usepackage{graphicx}
\usepackage{ dsfont }
\usepackage{booktabs}
\usepackage{longtable}

\usepackage[type={CC}, modifier={by-nc-sa}, version={4.0}]{doclicense}

\def\BibTeX{{\rm B\kern-.05em{\sc i\kern-.025em b}\kern-.08em
    T\kern-.1667em\lower.7ex\hbox{E}\kern-.125emX}}

\begin{document}

\title{\LARGE \bf
Data-driven Diffusion Models for Enhancing Safety in Autonomous Vehicle Traffic Simulations
}

\author{Jinxiong Lu$^{1}$, Shoaib Azam$^{1,2}$, Gokhan Alcan$^{3}$ and Ville Kyrki$^{1,2}$
\thanks{This work was supported by the Research Council of Finland Flagship Programme: Finnish Center for Artificial Intelligence (FCAI).}
\thanks{$^{1}$Jinxiong Lu, Shoaib Azam and Ville Kyrki are with the Department of Electrical Engineering and Automation, School of Electrical Engineering, Aalto University, Espoo 02150, Finland.}
\thanks{$^{2}$ Shoaib Azam and Ville Kyrki are also associated with Finnish Center for Artificial Intelligence.
{\tt\small \{jinxiong.lu; shoaib.azam; ville.kyrki\}@aalto.fi.}
{Corresponding Author: Shoaib Azam (shoaib.azam@aalto.fi)}}
\thanks{$^{3}$ Gokhan Alcan is with Automation Technology and Mechanical Engineering Unit at the Faculty of Engineering and Natural Sciences, Tampere University, Finland {\tt\small \{gokhan.alcan@tuni.fi\}.}}}%

\maketitle

\begin{abstract}
Safety-critical traffic scenarios are integral to the development and validation of autonomous driving systems. These scenarios provide crucial insights into vehicle responses under high-risk conditions rarely encountered in real-world settings. Recent advancements in critical scenario generation have demonstrated the superiority of diffusion-based approaches over traditional generative models in terms of effectiveness and realism. However, current diffusion-based methods fail to adequately address the complexity of driver behavior and traffic density information, both of which significantly influence driver decision-making processes. In this work, we present a novel approach to overcome these limitations by introducing adversarial guidance functions for diffusion models that incorporate behavior complexity and traffic density, thereby enhancing the generation of more effective and realistic safety-critical traffic scenarios. The proposed method is evaluated on two evaluation metrics: effectiveness and realism.The proposed method is evaluated on two evaluation metrics: effectiveness and realism, demonstrating better efficacy as compared to other state-of-the-art methods.


\end{abstract}

\section{INTRODUCTION}
The development of autonomous driving simulations allows systematic evaluation and continuous improvement of AD technologies, and a wide variety of challenging scenarios can be designed and used for testing. Within traffic simulations, safety-critical scenarios are the most important for improving safety in autonomous vehicles. The safety of autonomous vehicles is defined by the ability to navigate in near-collision (safety-critical) scenarios; however, these scenarios are rare in the real world, leading to data deficiency problems. In addition, as autonomous driving agents improve in terms of perception accuracy, decision-making capabilities, and control precision, these safety-critical scenarios occur even less frequently. This reduction in occurrence poses significant challenges for training autonomous systems, as the scarcity of high-risk situations makes it more difficult to expose agents to the edge cases that are essential for robust and reliable performance under critical conditions.
\par
Recent studies investigating the generation of safety-critical traffic scenarios have demonstrated that diffusion-based methods produce significantly more effective and realistic outcomes compared to the traditional generative models \cite{pronovost2023generating,saxena2021generative}. 
These probabilistic diffusion models, inspired by physical diffusion processes, have garnered significant attention due to their remarkable performance in computer vision and natural language processing tasks \cite{croitoru2023diffusion,yang2023diffusion}.  The data samples generated by these models exhibit not only high levels of realism but also considerable diversity. Notably, the application of diffusion models has recently expanded into the domain of autonomous driving, showcasing their potential to generate realistic traffic scenarios \cite{yang2024wcdt}.
\par
In safety-critical traffic scenario generation, near-collision driving behaviors are desired, such as tailgating, speeding, and overtaking. In diffusion model, guidance is a technique that steers the generation process to produce specific outcomes or samples conditioned on desired attributes, while still following the stochastic nature of the model. With guidance, safety-critical scenarios can be easily generated according to the specific guidance functions. These guidance functions can be trained value functions based on neural networks or heuristic guidance functions. 
\par
Although much advancement has been made in the field of incorporating adversarial guidance into diffusion probabilistic models, existing methods still have limitations \cite{chang2023controllable} \cite{xu2023diffscene}. Firstly, current approaches often overlook the behaviour complexity of human drivers, and the generated scenarios thus cannot fully capture the unpredictability of human drivers in real-world traffic. Secondly, the generated safety-critical traffic scenarios usually do not consider the influence of different traffic densities, even though traffic density information has a great effect on driver behaviour and decision-making. For example, in high-traffic density areas, vehicles need to stop and start more frequently, and react quickly to the behaviour of surrounding vehicles, such as lane merging and overtaking. In contrast, low-traffic density areas challenge the driver's ability to maintain a safe speed and be alert for sporadic vehicles passing by. Finally, existing approaches primarily focus on the speed constraint, but overlook the importance of constraining vehicles to be within road boundaries. All these limitations result in a gap in the applicability of the generated safety-critical traffic scenarios, as they cannot train and evaluate the AV system comprehensively across a wide variety of situations.
\par
Therefore, in this work, we proposed a novel diffusion-based method for safety-critical traffic scenarios generation with three innovative guidance functions. The proposed guidance functions are centered on behaviour complexity, traffic density, average speed. We evaluated our model on nuScence dataset and shows better performance in terms of realism and effectiveness as compared to state-of-the-art method.

The main contributions of our work are:
\begin{enumerate}
    \item We proposed three novel guidance functions for diffusion probabilistic models to overcome the limitations in safety-critical traffic scenarios generation.
    \item We experimentally evaluated each guidance functions to the effectiveness and realism of the generated scenarios.
\end{enumerate}


\section{Related Work}
In this section, we provide an overview of previous work related to safety-critical traffic scenario generation.

While general and controllable traffic scenario generations are crucial, the ability to generate and evaluate autonomous vehicles against safety-critical scenarios is arguably the most vital. Safety-critical traffic scenarios are designed to produce conditions that are likely to challenge the vehicle's planning and decision-making capabilities to their limits. These scenarios are particularly valuable because they simulate rare but potentially catastrophic situations that AVs must be prepared to handle.

One of the prominent advancement in safety-critical scenario generation was STRIVE \cite{rempe2022generating}, which was a novel method aiming to generate both challenging and realistic traffic scenarios. It uses a graph-based conditional Variational Autoencoder to simulate scenarios where AV systems are likely to fail, and it defines the traffic scenario generation problem as an optimization problem in the latent space. Although the generated traffic scenarios are quite challenging, their realism does not deteriorate significantly. 

While previous work like STRIVE has laid the groundwork for safety-critical scenario generation by optimizing latent spaces, the advancement of diffusion-based models for safety-critical traffic simulation through adversarial guidance creates a new direction for generating realistic and challenging traffic scenarios for autonomous vehicles training and testing, particularly through the work of \cite{xu2023diffscene} and \cite{chang2023controllable}.

DiffScene \cite{xu2023diffscene} is a diffusion probabilistic model-based safety-critical scenario generation method, representing a significant advancement compared with previous methods such as STRIVE. It trains the diffusion model with a huge amount of data to ensure the realism of the generated trajectories and applies adversarial objectives in test-time guidance to ensure the generation of safety-critical scenarios. These adversarial objectives can be divided into three categories: safety, functionality, and constraint objectives. DiffScene was evaluated in terms of effectiveness and realism, and the effectiveness metrics include collision rate, route incompletion rate, and speed satisfaction. Finally, the evaluation results show that DiffScene outperformed contemporary state-of-the-art approaches both in effectiveness and realism. 

Another prominent advancement in safety-critical scenario generation based on the diffusion model can be found in Chang's work \cite{chang2023controllable}. It proposed a few novel adversarial guidance functions aiming to improve long-timestep traffic simulation and achieve a balance between adversarial challenges and realism. These adversarial functions include collision distance, time to collision, lane margin, and relative speed objectives. One noticeable difference compared with DiffScene is that it adds an extra type of vehicle, namely non-adversarial vehicles, in addition to the ego vehicle and the adversarial vehicle. 

Our work compliments previous work with three innovative guidance functions centered on behaviour complexity, traffic density, average speed

\section{Method}
This section explains our proposed method for safety-critical scenarios generation. First, we addresses the problem formulation of safety critical traffic generation, followed by diffusion-based scene generation, and finally introduce the adversarial optimization of the guided sampling process.

\subsection{Problem formulation}
\label{Problem}

 \begin{figure*}[h]
  \centering
  \includegraphics[width=0.6\textwidth]{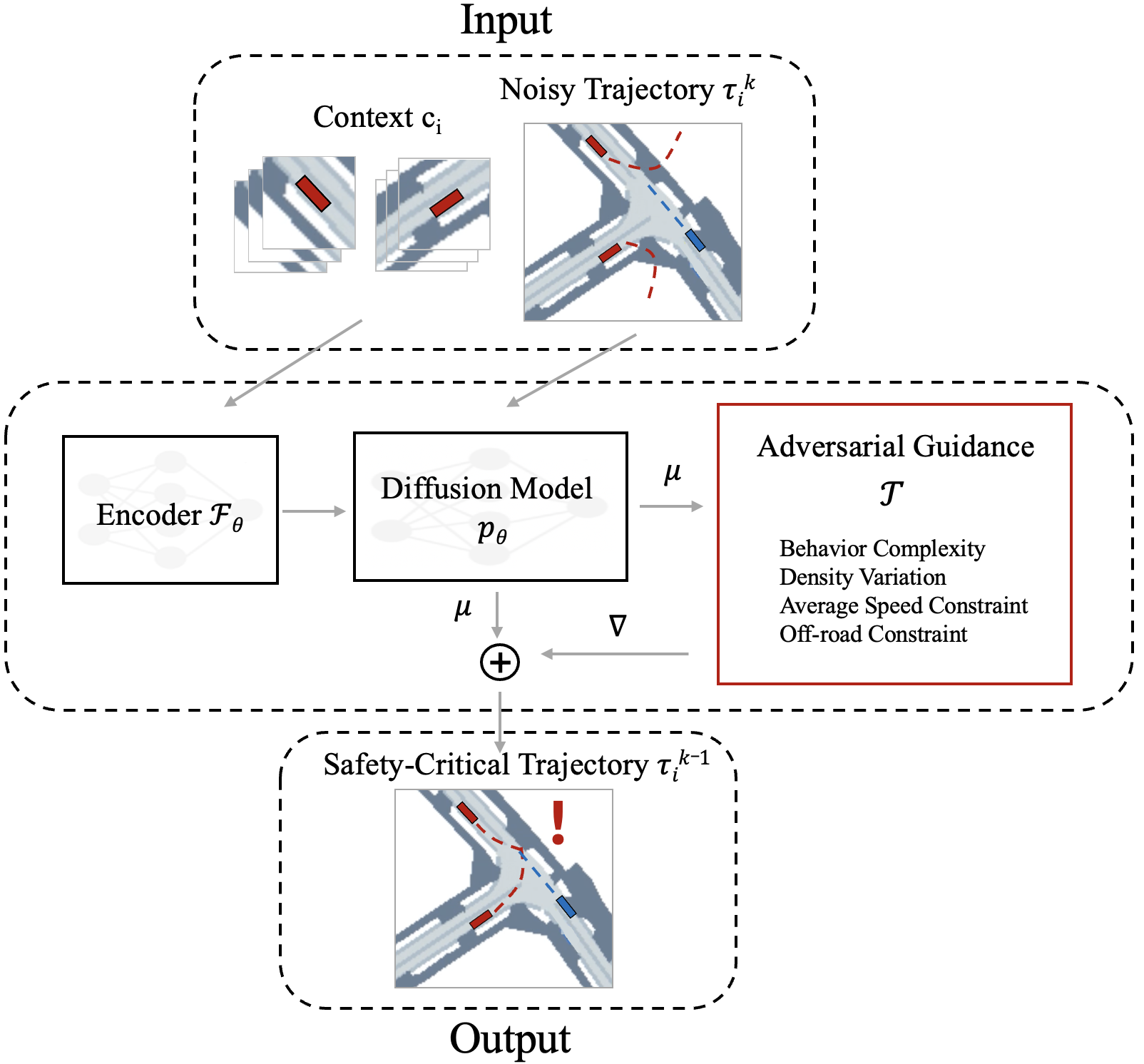}
  \caption{Guided Diffusion Process for the Adversarial Agent (red vehicles are adversarial agents, the blue vehicle is the ego vehicle controlled by a deep reinforcement learning algorithm, i is the index of an adversarial vehicle, k is the diffusion step).}
  \label{fig:GOA}
\end{figure*}

The overview of the proposed approach and an example of the generated traffic scenario are presented in Figure \ref{fig:GOA}. Context information of each safety-critical vehicle is encoded before feeding to the diffusion model, and the gradient from adversarial guidance is required to modify the generated trajectory to be more safety-critical.

The target vehicle's state at timestep $t$, denoted as $s_{t}^{\text{tgt}}$, is represented by a four-dimensional vector $(x_{t}^{\text{tgt}}, y_{t}^{\text{tgt}}, v_{t}^{\text{tgt}}, \theta_{t}^{\text{tgt}})$, encompassing its two-dimensional position, longitudinal velocity, and yaw angle respectively.Control variables for the safety-critical vehicle $a_{t}^{\text{tgt}}$ include acceleration and yaw velocity $(\dot{v}_{t}^{\text{tgt}}, \dot{\theta}_{t}^{\text{tgt}})$. The context $\mathbf{c}=(I, S)$ for the encoder includes a agent-centric semantic map $I$ and the $H$ previous states of both the target agent and its $M$ neighbors $S_{t-H:t} = \{s_{t-H:t}^{\text{tgt}}, s_{t-H:t}^{1}, \cdots, s_{t-H:t}^{M}\}$. To calculate the state of vehicle $m$ at timestep $t+1$, a transition function $\mathcal{D}$ is needed (unicycle dynamics model is used in this study), and it can calculate the state $s_{t+1}^{m}$ based on the preceding state $s_{t}^{m}$ and the control variables $a_{t}^{m}$, with $s_{t+1}^{m} = \mathcal{D}(s_{t}^{m}, a_{t}^{m})$. 

In the safety-critical scenarios generation, traffic participants include $\mathcal{U} := \{u_{\text{ego}}, u_{\text{sv1}}, \ u_{\text{sv2}}, \cdots, u_{\text{svn}} \}$, where $u_{\text{ego}}$ is the ego vehicle controlled by a deep reinforcement learning algorithm $f$ and $ \{ u_{\text{sv1}}, u_{\text{sv2}}, \cdots, u_{\text{svn}} \}$ are safety-critical vehicles controlled by the guided diffusion model.

Let the state and action trajectory of a target vehicle from timestep $0$ to $T$ be $\tau_{s}^{tgt} = \{s^{tgt}_{0}, s^{tgt}_{1}, \cdots, s^{tgt}_{T}\}$ and $\tau_{a}^{tgt} = \{a^{tgt}_{0}, a^{tgt}_{1}, \cdots, a^{tgt}_{T}\}$ . The state trajectory of a safety-critical vehicle can be denoted as $\tau_{s}^{\text{sv}}=\{\mathbf{s}^{sv1}, \mathbf{s}^{sv2}, \cdots, \mathbf{s}^{svn}\}$.

The objective of the safety-critical scenario generation is to generate a trajectory $\tau_{s}^{\text{sv}}$ for each safety-critical vehicle, which maximizes the cost function measuring the risk of the current scenario and the satisfaction of constraints $\mathcal{R}_{\text{adv}}(f,\tau_{s}^{\text{sv}})$ while keeping the value of another cost function measuring the realism of the current scenario $N(\tau_{s}^{\text{sv}})$ within certain threshold:

\begin{equation}
\arg \max_{\tau^{\text{sv}}} \mathcal{R}_{\text{adv}}(f,\tau_{s}^{\text{sv}}), \quad \text{s.t. } N(\tau_{s}^{\text{sv}}) < c, \quad 
\end{equation}
where $c$ is a threshold used in the realism cost function.

To achieve the mentioned objective, the diffusion model is first trained on large-scale data set to generate general traffic scenarios with small realism loss. Then adversarial optimization is applied in the guided sampling process to maximize the driving risk for the ego vehicle while satisfying certain constraints.

\subsection{Conditional Diffusion for Traffic Modeling}
\label{Conditional}
Diffusion models treat sample generation as an iterative process of denoising through the learning of a reverse diffusion process. The trajectories generated by the diffusion model include state and action trajectories, and they are generated by iteratively deducting the noise in the trajectories starting from pure Gaussian noise. This diffusion model is conditional, as its denoising process is conditioned on the context information encoded by the ResNet-based encoder.

\subsubsection{Trajectory representation}
The trajectories generated by the diffusion model can be denoted as:

\begin{equation}
\tau := \begin{bmatrix}
\tau_a \\
\tau_s
\end{bmatrix}
\end{equation}

Instead of jointly predicting sate and action trajectories \cite{janner2022planning}, the proposed method only predicts action trajectories $\tau_a$ and utilizes known dynamic model $\mathcal{D}$ and initial state $s_0$ to infer the state trajectories $\tau_s$, denoted as $\tau_s = \mathcal{M}(s_0, \tau_a)$, which ensure the physical feasibility of the state trajectories during denoising.

\subsubsection{Formulation}
Let $\tau_a^k$ be the action trajectory in the $k^{th}$ diffusion step, where $k = 0$ corresponds to the clean trajectory. Diffusion process acted on $\tau_a^0$ can be expressed as: 

\begin{equation}
q(\tau_a^{1:K}|\tau_a^0) := \prod_{k=1}^{K} q(\tau_a^k|\tau_a^{k-1}) \quad
\end{equation}

\begin{equation}
q(\tau_a^k|\tau_a^{k-1}) := \mathcal{N}(\tau_a^k; \sqrt{1 - \beta_k} \tau_a^{k-1}, \beta_k \mathbf{I}),
\end{equation}
where $\beta_1, \beta_2, \cdots, \beta_K$ are variance that controls the amount of the Gaussian noise added to the trajectory at every diffusion step. In the limit of infinite diffusion steps, the signal is corrupted
into an isotropic Gaussian distribution.

For trajectory generation, the trained UNet-based model is applied to reverse the diffusion process, which iteratively denoises the noisy trajectory \cite{janner2022planning}. The reverse diffusion process can be expressed as:

\begin{equation}
p_{\theta}(\tau_a^{0:K}|\mathbf{c}) := p(\tau_a^K) \prod_{k=1}^{K} p_{\theta}(\tau_a^{k-1}|\tau_a^k, \mathbf{c})
\end{equation}

\begin{equation}
p_{\theta}(\tau_a^{k-1}|\tau_a^k, \mathbf{c}) := \mathcal{N}(\tau_a^{k-1}; \mu_{\theta}(\tau_a^k, k, \mathbf{c}), \Sigma_{\theta}(\tau_a^k, k, \mathbf{c})), \quad
\end{equation}
where $\mathbf{c}$ is the environment context information and $p(\tau_a^K) = \mathcal{N}(0, \mathbf{I})$,$\theta$ are parameters for the diffusion model. The input of the model includes both the action trajectories $\tau_a^k$ and the resulting states trajectories $\tau_s^k = \mathcal{M}(s_0, \tau_a^k)$. Following \cite{gu2022stochastic}, the variance of the Gaussian transition is set as $\Sigma^k = \Sigma_{\theta}(\tau_a^k, k, \mathbf{c}) = \sigma_{\theta}^2 \mathbf{I} = \beta_k \mathbf{I}$.

\subsubsection{Training}

In every training timestep, the environment context information $\mathbf{c}$ and ground truth clean trajectory $\tau^0$ are sampled from a real-world driving dataset and the denoising step $k$ is uniformly sampled from $\{1, \ldots, K\}$. The action trajectories are first corrupted by computing the noisy trajectories $\tau^k$ from the clean trajectories $\tau^0$ with $\tau_a^k = \sqrt{\bar{\alpha}_k} \tau_a^0 + \sqrt{1 - \bar{\alpha}_k} \epsilon$, where $\epsilon \sim \mathcal{N}(0, \mathbf{I})$ and $\bar{\alpha}_k = \prod_{l=0}^{k} 1 - \beta_l$, and then computing the corresponding state $\tau_s^k = \mathcal{M}(s_0, \tau_a^k)$.

The diffusion model indirectly parameterizes $\mu_{\theta}$ in Eq. \ref{eq:loss} by instead predicting the uncorrupted trajectory $\hat{\tau}^0 = [\hat{\tau}_a^0; f(s_0, \hat{\tau}_a^0)]$ where $\hat{\tau}_a^0 = \hat{\tau}_a^0(\tau_a^k, k, \mathbf{c})$ is the direct network output (see \cite{li2022diffusion}, \cite{nichol2021improved}). Finally, the loss function used to train the model is:

\begin{equation}
\label{eq:loss}
L(\theta) = \mathbb{E}_{\epsilon, k, \tau^0, \mathbf{c}} \left[\left\| \tau^0 - \hat{\tau}^0 \right\|^2\right]. \quad
\end{equation}

Both action and state trajectories are trained with this loss function, and previous results have shown that the addition of state trajectories increases the generation quality of the traffic scenarios \cite{janner2022planning}.

\subsection{Guided Adversarial Optimization}
\label{Guided}
Although training the diffusion model on a large-scale real-world driving dataset can generate realistic trajectories that achieve low realism loss, it still lacks the properties of being safety-critical. Therefore, introducing guided adversarial optimization into the guided sampling process is necessary to generate safety-critical and realistic traffic scenarios.

In the reverse diffusion process, an adversarial function $\mathcal{J}(\tau)$ is defined to guide the process toward objectives specified by the adversarial function. At every reverse diffusion timestep, the gradient of the adversarial function $\mathcal{J}$ is used to modify the denoising process:

\begin{equation}
p_{\theta}(\tau^{k-1}|\tau^k) \approx \mathcal{N}(\tau^{k-1}; \mu + \Sigma g, \Sigma), \quad (11)
\end{equation}
where $\mu$ and $\Sigma$ are the mean and variance of the Gaussian transition, $g = \nabla \mathcal{J}(\tau)$ specifies the optimization direction. Through iteratively optimizing the vehicle trajectory toward better satisfaction of $\mathcal{J}$, the diffusion model can eventually generate effective and realistic safety-critical traffic scenarios. 

This adversarial optimization process offers flexibility to the generation of traffic simulation since the adversarial guidance functions can be designed and tailored to meet users' specific needs. In this study, two types of objectives with innovative guidance functions are introduced: functionality-based objectives and constraint-based objectives. Firstly, functionality-based objectives include $\mathcal{J}_{\text{BC}}(\tau)$, which focuses on the behaviour complexity of the SV, and $\mathcal{J}_{\text{DV}}(\tau)$, which focuses on the traffic density of the SV. Secondly, constraint-based objectives include $\mathcal{J}_{\text{AS}}(\tau)$, which focuses on the average speed of the SV, and $\mathcal{J}_{\text{OR}}(\tau)$ \cite{zhong2023guided}, which focuses on keeping SV within the road boundary.

Finally, the adversarial guidance function aimed to be maximized in this study is a combination of four innovative guidance functions:

\begin{equation}
\mathcal{J}(\tau) = \omega_{b} \mathcal{J}_{\text{BC}}(\tau) + \omega_{d} \mathcal{J}_{\text{DV}}(\tau) + \omega_{a} \mathcal{J}_{\text{AS}}(\tau) + \omega_{o} \mathcal{J}_{\text{OR}}(\tau) \quad 
\end{equation}
where $\omega_b$, $\omega_d$, $\omega_a$, and $\omega_o$ are hyperparameters controlling the weights of four different objectives.

\subsubsection{Behaviour Complexity}

Behaviour complexity guidance aims to promote the diversity of driving characteristics. Existing diffusion-based methods overlook driver behavior complexity, even though the unpredictability of other human drivers has a great effect on driver decision-making. Behavior complexity guidance is modeled as:

\begin{equation}
\mathcal{J}_{\text{BC}}(\tau) = \sum_{t=0}^{T} (w_{acc} |\Delta v_t| + w_{\theta} | \Delta \theta|)
\end{equation}
where $\Delta v_t$ represents the change in speed from one timestep to the next, i.e. acceleration. $\Delta \theta$ represents the change in direction (yaw) from one timestep to the next, i.e. rotational velocity. $w_{acc}$ and $w_{\theta}$ are the corresponding weights for $|\Delta v_t|$ and $|\Delta \theta|$.

\subsubsection{Traffic Density}

Traffic density guidance aims to keep the traffic density close to a target value. Depending on the requirements, the target vehicle density can be set to relatively high or low. For example, in high-density traffic, vehicles need to make more frequent stops and starts and react quickly to sudden changes, such as a nearby vehicle abruptly changing lanes. In contrast, lower traffic densities might challenge the vehicle’s ability to maintain safe speeds and remain alert for sporadic obstacles or vehicles merging lanes. In safety-critical traffic scenario generation, higher traffic density is more desired. Traffic density guidance is modeled as:

\begin{equation}
\mathcal{J}_{\text{DV}}(\tau) = - \sum_{t=0}^{T} |\rho_t - \rho_{d}|
\end{equation}
where vehicle density $\rho_t$ at any given time $t$ is defined as the number of vehicles per unit area within a specific region surrounding the autonomous vehicle. Target vehicle density $\rho_{d}$ is the desired traffic density. Region of Interest (ROI) could be a defined radius around the AV or a specific area on the map where traffic density is being monitored. When implementing traffic density guidance, a differentiable indicator function
to represent if the target vehicle is within the region of interest is required. Common
choices of differentiable indicator functions include the sigmoid function and the tanh
function.

In this study, the ROI is selected as a circular region centered on the ego vehicle, with a radius parameter to control the size of the ROI. The target number of vehicle is set to be 5 throughout this study and thus the target vehicle density can be derived from the ROI radius.

\subsubsection{Average speed Constraint}

Average speed guidance aims to keep the average speed of SV close to a desired value. Instead of controlling the speed of SV at every timestep as in \cite{xu2023diffscene}, this study chose to constraint the average speed in order to facilitate behaviour complexity. The main advantage of average speed guidance compared with Target Speed guidance is that it can regulate the speeds of safety-critical vehicles to be within the normal driving speed range while not sacrificing behaviour diversity. For example, when the current speed of a safety-critical vehicle is exactly the target speed, Target Speed guidance will try to set the acceleration to zeros, stifling behaviour diversity, whereas average speed guidance only regulates the average speed across the complete trajectory instead of focusing on the individual speed at every timestep. Average speed guidance is modeled as:

\begin{equation}
    \mathcal{J}_{\text{AS}}(\tau) = - |\frac{\sum_{t=0}^{T} v_t }{T} - v_d|
\end{equation}
where $v_d$ is the desired average speed of SV.

\subsubsection{Off-road Constraint}

Off-road Constraint aims to keep the vehicles within the boundaries of the road. Off-road Constraint is important because it can regularize the behaviour complexity guidance. Applying behaviour complexity guidance alone will lead SVs to drive out of road boundaries frequently. Thus, the combination of behaviour complexity guidance and Off-road constraint guidance is essential and denoted as augmented behaviour complexity (AB) guidance. Off-road guidance is modeled as:

\begin{equation}
    \mathcal{J}_{\text{OR}}(\tau) = \sum_{t=0}^{T} D_{t}
\end{equation}
where $D_{t}$ denotes the distance from the vehicle center to the closest point that is on the road boundary at time $t$ \cite{zhong2023guided}. When the center of the vehicle is within the road boundary, $D_{t}$ is positive; otherwise, it is negative.

\section{Experiments}
In this section, we describe the evaluation metrics and evaluate the proposed method with respect to a state-of-the-art baseline DiffScene \cite{xu2023diffscene}. Finally, we conduct an ablation study to identify the contribution of each guidance functions and make sure there are no redundant elements that do not contribute to the effectiveness and realism.

\subsection{Evaluation metrics}

\subsubsection{Effectiveness}
An effective safety-critical traffic simulation should keep challenging the ego vehicle while satisfying physical constraints. Therefore, the effectiveness of the generated traffic scenarios is evaluated with the following three metrics: Collision Rate (CR) of the ego vehicle, which can be computed as $\mathbb{E}_{\tau \sim \mathcal{P}}[\mathds{1}(\tau)]$, where $\mathcal{P}$ is the distribution of the generated trajectory. Route Incompletion Rate (IR) of the ego vehicle, which can be computed as $\mathbb{E}_{\tau \sim \mathcal{P}}[r(\tau)]$. Speed Satisfaction (SS) of the safety-critical vehicles, which can be computed as $\mathbb{E}_{\tau \sim \mathcal{P}}[\mathbb{E}_t[(1-\frac{|v_t - v_d|}{v_d})]]$, where $\mathds{1}$ is an indicator function for collision, $v_d$ is the desired speed. According to the SafeBench benchmarking platform \cite{xu2022safebench}, the desired velocity is set to 8, which is usually the normal driving speed of vehicles.

\subsubsection{Realism}
To evaluate the realism of the generated traffic simulation, the trajectories from the dataset and the generated simulation are compared. This comparison is achieved by computing the Wasserstein distance between the normalized histograms of the driving properties of the simulated and dataset trajectories. Inspired by Pavone's work \cite{zhong2023guided}, the evaluated driving properties include longitudinal acceleration magnitude, latitudinal acceleration magnitude, and total jerk, and this metric is named as realism deviation.

\subsection{Comparison with a state-of-the-art algorithm}
\label{sota_compare}
\begin{table}[h!]
    \centering
    \setlength{\tabcolsep}{12pt} 
    \renewcommand{\arraystretch}{1.4} 
    \begin{tabular}{c|cccc}
        \toprule
        & \textbf{CR $\uparrow$} & \textbf{IR $\uparrow$} & \textbf{SS $\uparrow$} & \textbf{RD $\downarrow$} \\ 
        \midrule
        \textbf{DiffScene} & 0.58 & 0.61 & \textbf{0.41} & 0.40 \\ 
        \textbf{Ours} & \textbf{0.64} & \textbf{0.69} & 0.35 & \textbf{0.36} \\
        \bottomrule
    \end{tabular}
    \caption{Comparison with the state-of-the-art safety-critical traffic scenario generator, DiffScene. (CR: collision rate, IR: route incompletion rate, SS: speed satisfaction, RD: realism deviation. Arrows indicate the direction of better results.)}
    \label{tab:SOTA}
\end{table}


A comparison with a state-of-the-art safety-critical traffic scenarios generator DiffScene, is shown in Table \ref{tab:SOTA}. Except for the speed satisfaction metrics, the proposed method achieves better performance in all other metrics, including collision rate, route incompletion rate, and realism deviation. It is difficult to outperform DiffScene in terms of speed satisfaction metrics because DiffScene uses target speed guidance in the adversarial optimization process, which directly influences the speed at every timestep to be the exact same target speed used in the speed satisfaction metrics. Even though an average speed guidance is applied in the proposed method, which only constrains the average speed of the safety-critical vehicles across the complete trajectory timestep, the proposed method maintains a fairly high value of speed satisfaction, only $5.5\%$ smaller than DiffScene. 

However, when it comes to other metrics, the proposed method outperforms DiffScene significantly, with a $64\%$ collision rate and $69\%$ route incompletion rate. Compared with DiffScene, the proposed method is able to generate more challenging safety-critical traffic scenarios, while achieving higher realism (lower realism deviation). 

\subsection{Ablation Study}
\label{ablation}
\begin{table}[h!]
    \centering
    \caption{Ablation study of the proposed guidance functions. (AB: augmented behavior complexity guidance, DV: traffic density guidance, AS: average speed guidance, CR: collision rate, IR: route incompletion rate, SS: speed satisfaction, RD: realism deviation. Arrows indicate the direction of better results.)}
    \label{tab:Ablation}
    \setlength{\tabcolsep}{10pt} 
    \renewcommand{\arraystretch}{1.4} 
    \begin{tabular}{ccc|cccc}
        \toprule
        \textbf{AB} & \textbf{DV} & \textbf{AS} & \textbf{CR $\uparrow$} & \textbf{IR $\uparrow$} & \textbf{SS $\uparrow$} & \textbf{RD $\downarrow$} \\
        \midrule
        $\checkmark$ & $\checkmark$ & $\checkmark$ & \textbf{0.64} & \textbf{0.69} & 0.35  & 0.36 \\
        $\checkmark$ & $\checkmark$ & $\times$     & 0.62 & 0.68 & 0.26  & 0.39 \\
        $\times$     & $\checkmark$ & $\checkmark$ & 0.31 & 0.53 & \textbf{0.36} & 0.37 \\
        $\checkmark$ & $\times$     & $\checkmark$ & 0.62 & 0.68 & 0.35  & 0.36 \\
        $\checkmark$ & $\times$     & $\times$     & 0.38 & 0.59 & 0.24  & 0.38 \\
        $\times$     & $\checkmark$ & $\times$     & 0.54 & 0.59 & 0.26  & \textbf{0.35} \\
        $\times$     & $\times$     & $\checkmark$ & 0.38 & 0.57 & \textbf{0.36} & 0.37 \\
        $\times$     & $\times$     & $\times$     & 0.23 & 0.50 & 0.28  & \textbf{0.35} \\
        \bottomrule
    \end{tabular}
\end{table}
Ablation study is conducted in order to identify the contribution of each guidance functions and make sure there are no redundant elements that do not contribute to the effectiveness and realism. The ablation study of the proposed guidance functions is shown in Table \ref{tab:Ablation}, with the best results highlighted in bold.

First, the influence of each individual guidance on the baseline is analyzed. Using augmented behavior complexity guidance alone increases collision rate, route incompletion, and realism deviation, while speed satisfaction drops to 0.24, indicating higher risk with minor realism loss. Traffic density guidance shows similar effects but performs better overall, with a lower collision rate (0.38 to 0.54), better speed satisfaction, and no deterioration in realism. Average speed guidance significantly improves speed satisfaction, with moderate increases in collision and route incompletion rates, and only slight realism deterioration.

Secondly, three combinations of two guidance functions are analyzed. Combining behavior complexity and traffic density yields better results in collision rate, route incompletion, and speed satisfaction, but with the highest realism deviation. Pairing behavior complexity with average speed offers similar collision and route results but with better speed satisfaction and realism. Combining traffic density and average speed achieves the best speed satisfaction but underperforms in other metrics.

Lastly, the proposed method with three guidance functions is analyzed. Compared with the baseline, the proposed approach improved significantly in Collision rate, route incompletion rate, and speed satisfaction metrics, and only lost a small amount of realism, with realism deviation rising from $0.35$ to $0.36$. This result indicates that the proposed guidance function is able to more dangerous safety-critical traffic scenarios with only a minor loss in realism.

\section{Conclusion}

To address the challenge of generating diverse and effective safety-critical traffic scenarios for the comprehensive training and evaluation of autonomous vehicles, this work incorporated innovative guidance functions into the diffusion model,  overcoming the limitations of existing heuristic-based traffic simulation and state-of-the-art diffusion-based safety-critical traffic scenario generators. Then the proposed approach was evaluated in three aspects, including safety, functionality, and constraint, achieving better performance than a state-of-the-art algorithm both in terms of effectiveness and realism. In addition, an ablation study on the proposed adversarial guidance function was conducted, demonstrating that each guidance function significantly contributed to the effectiveness and realism of the generated safety-critical scenarios.

This work highlights the potential of integrating adversarial guidance into probabilistic diffusion models. While effective in enhancing realism and safety in traffic simulations, broader applications remain unexplored. Comprehensive studies on applying adversarial guidance to diverse traffic participants are lacking. Future research could extend these methods to include pedestrians, cyclists, and emergency vehicles, advancing safer and more adaptive autonomous driving systems.

\bibliographystyle{IEEEtran}
\bibliography{ref}

\begin{thebibliography}{10}
\providecommand{\url}[1]{#1}
\csname url@samestyle\endcsname
\providecommand{\newblock}{\relax}
\providecommand{\bibinfo}[2]{#2}
\providecommand{\BIBentrySTDinterwordspacing}{\spaceskip=0pt\relax}
\providecommand{\BIBentryALTinterwordstretchfactor}{4}
\providecommand{\BIBentryALTinterwordspacing}{\spaceskip=\fontdimen2\font plus
\BIBentryALTinterwordstretchfactor\fontdimen3\font minus
  \fontdimen4\font\relax}
\providecommand{\BIBforeignlanguage}[2]{{%
\expandafter\ifx\csname l@#1\endcsname\relax
\typeout{** WARNING: IEEEtran.bst: No hyphenation pattern has been}%
\typeout{** loaded for the language `#1'. Using the pattern for}%
\typeout{** the default language instead.}%
\else
\language=\csname l@#1\endcsname
\fi
#2}}
\providecommand{\BIBdecl}{\relax}
\BIBdecl

\bibitem{pronovost2023generating}
E.~Pronovost, K.~Wang, and N.~Roy, ``Generating driving scenes with
  diffusion,'' \emph{arXiv preprint arXiv:2305.18452}, 2023.

\bibitem{saxena2021generative}
D.~Saxena and J.~Cao, ``Generative adversarial networks (gans) challenges,
  solutions, and future directions,'' \emph{ACM Computing Surveys (CSUR)},
  vol.~54, no.~3, pp. 1--42, 2021.

\bibitem{croitoru2023diffusion}
F.-A. Croitoru, V.~Hondru, R.~T. Ionescu, and M.~Shah, ``Diffusion models in
  vision: A survey,'' \emph{IEEE Transactions on Pattern Analysis and Machine
  Intelligence}, vol.~45, no.~9, pp. 10\,850--10\,869, 2023.

\bibitem{yang2023diffusion}
L.~Yang, Z.~Zhang, Y.~Song, S.~Hong, R.~Xu, Y.~Zhao, W.~Zhang, B.~Cui, and
  M.-H. Yang, ``Diffusion models: A comprehensive survey of methods and
  applications,'' \emph{ACM Computing Surveys}, vol.~56, no.~4, pp. 1--39,
  2023.

\bibitem{yang2024wcdt}
C.~Yang, A.~X. Tian, D.~Chen, T.~Shi, and A.~Heydarian, ``Wcdt: World-centric
  diffusion transformer for traffic scene generation,'' \emph{arXiv preprint
  arXiv:2404.02082}, 2024.

\bibitem{chang2023controllable}
W.-J. Chang, F.~Pittaluga, M.~Tomizuka, W.~Zhan, and M.~Chandraker,
  ``Controllable safety-critical closed-loop traffic simulation via guided
  diffusion,'' \emph{arXiv preprint arXiv:2401.00391}, 2023.

\bibitem{xu2023diffscene}
C.~Xu, D.~Zhao, A.~Sangiovanni-Vincentelli, and B.~Li, ``Diffscene:
  Diffusion-based safety-critical scenario generation for autonomous
  vehicles,'' in \emph{The Second Workshop on New Frontiers in Adversarial
  Machine Learning}, 2023.

\bibitem{rempe2022generating}
D.~Rempe, J.~Philion, L.~J. Guibas, S.~Fidler, and O.~Litany, ``Generating
  useful accident-prone driving scenarios via a learned traffic prior,'' in
  \emph{Proceedings of the IEEE/CVF Conference on Computer Vision and Pattern
  Recognition}, 2022, pp. 17\,305--17\,315.

\bibitem{janner2022planning}
M.~Janner, Y.~Du, J.~B. Tenenbaum, and S.~Levine, ``Planning with diffusion for
  flexible behavior synthesis,'' \emph{arXiv preprint arXiv:2205.09991}, 2022.

\bibitem{gu2022stochastic}
T.~Gu, G.~Chen, J.~Li, C.~Lin, Y.~Rao, J.~Zhou, and J.~Lu, ``Stochastic
  trajectory prediction via motion indeterminacy diffusion,'' in
  \emph{Proceedings of the IEEE/CVF Conference on Computer Vision and Pattern
  Recognition}, 2022, pp. 17\,113--17\,122.

\bibitem{li2022diffusion}
X.~Li, J.~Thickstun, I.~Gulrajani, P.~S. Liang, and T.~B. Hashimoto,
  ``Diffusion-lm improves controllable text generation,'' \emph{Advances in
  Neural Information Processing Systems}, vol.~35, pp. 4328--4343, 2022.

\bibitem{nichol2021improved}
A.~Q. Nichol and P.~Dhariwal, ``Improved denoising diffusion probabilistic
  models,'' in \emph{International conference on machine learning}.\hskip 1em
  plus 0.5em minus 0.4em\relax PMLR, 2021, pp. 8162--8171.

\bibitem{zhong2023guided}
Z.~Zhong, D.~Rempe, D.~Xu, Y.~Chen, S.~Veer, T.~Che, B.~Ray, and M.~Pavone,
  ``Guided conditional diffusion for controllable traffic simulation,'' in
  \emph{2023 IEEE International Conference on Robotics and Automation
  (ICRA)}.\hskip 1em plus 0.5em minus 0.4em\relax IEEE, 2023, pp. 3560--3566.

\bibitem{xu2022safebench}
C.~Xu, W.~Ding, W.~Lyu, Z.~Liu, S.~Wang, Y.~He, H.~Hu, D.~Zhao, and B.~Li,
  ``Safebench: A benchmarking platform for safety evaluation of autonomous
  vehicles,'' \emph{Advances in Neural Information Processing Systems},
  vol.~35, pp. 25\,667--25\,682, 2022.

\end{thebibliography}

\end{document}